%% file: main.tex
\documentclass{article}

\PassOptionsToPackage{numbers, compress}{natbib}

\usepackage[preprint]{neurips_2026}

\usepackage[utf8]{inputenc} %
\usepackage[T1]{fontenc}    %
\usepackage{makecell}
\usepackage{booktabs} %
\usepackage[colorlinks, citecolor=green, linkcolor=blue, urlcolor=blue]{hyperref}

\usepackage{url}            %
\usepackage{amsfonts}       %
\usepackage{nicefrac}       %
\usepackage{microtype}      %
\usepackage[table]{xcolor}  %

\input{macros.tex}          %

\title{DyaPlex: Full-Duplex Speech-Motion Model for Dyadic Interaction}

\author{
\textbf{Koki Nagano\textsuperscript{\rm 1}\footnotemark[2]}, \
\textbf{Hongyu Liu\textsuperscript{\rm 1,2}\thanks{Part of the work was done during an internship at NVIDIA.}\kern0.4em\thanks{Joint first authors.}~}, \
\textbf{Seonwook Park\textsuperscript{\rm 1}}, \
\textbf{Tianye Li\textsuperscript{\rm 1}} \\
\textbf{Amrita Mazumdar\textsuperscript{\rm 1}}, \
\textbf{Christian Jacobsen\textsuperscript{\rm 1}}, \
\textbf{Shengze Wang\textsuperscript{\rm 1}}, \
\textbf{Michael Stengel\textsuperscript{\rm 1}} \\
\textbf{Rajarshi Roy\textsuperscript{\rm 1}}, \
\textbf{Ka Chun Cheung\textsuperscript{\rm 1}}, \
\textbf{Simon See\textsuperscript{\rm 1}}, \
\textbf{Shalini De Mello\textsuperscript{\rm 1}} \\
\textsuperscript{\rm 1}NVIDIA, \
\textsuperscript{\rm 2}HKUST
}

\begin{document}

\maketitle

\begin{center}
  \includegraphics[width=\linewidth]{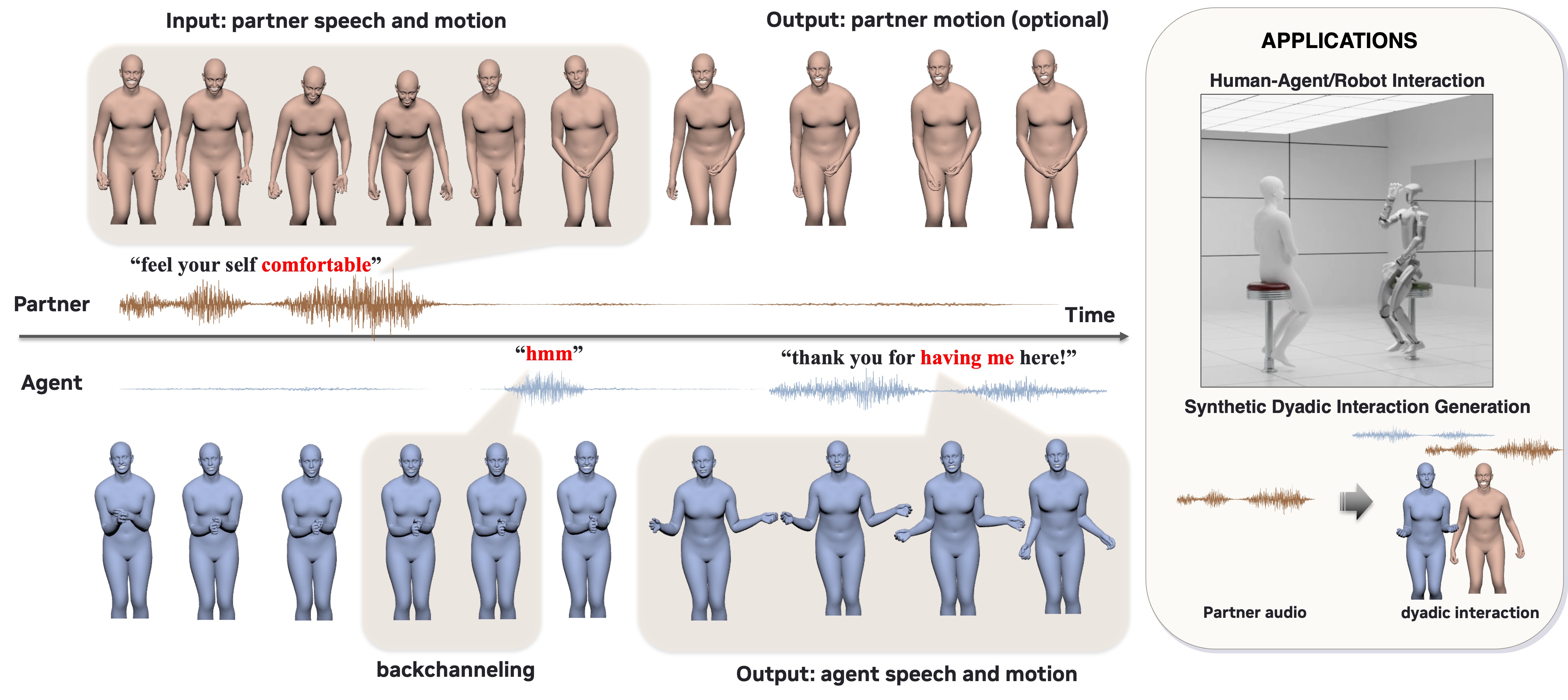}
  \captionof{figure}{\textbf{DyaPlex} is a causal, full-duplex speech-motion model that simultaneously speaks and listens to a partner while perceiving a partner motion and generating agent's motion. Our model could be applied to applications, such as, dyadic interactions with a human user and agent/robot (right) as well as generating synthetic speech-motion dyadic interaction data. 
  }
  
  \label{fig:teaser}
  \end{center}

\begin{abstract}

We present DyaPlex, a streaming, full-duplex speech-and-motion model designed for  dyadic interaction. To capture the continuous and reciprocal nature of human communication, this full-duplex capability empowers the agent to simultaneously perceive and generate both speech and physical motion in a streaming fashion. At its core, our method leverages the strong priors of a foundational full-duplex speech model and integrates a novel motion pathway, thereby achieving fully synchronized multi-modal interaction. Specifically, we design a dual-tower Transformer architecture that preserves the zero-shot conversational reasoning of a frozen base speech model while constructing a deeply coupled, streaming motion pathway. By introducing a unified dyadic token interleaving mechanism and guiding cross-attention via a time-aligned speech-motion RoPE, our model effectively aligns autoregressive motions with rich latent speech features. Trained on the 4,000-hour Seamless Interaction dataset, our model effectively captures cross-speaker dependencies and establishes new state-of-the-art performance across both monadic and dyadic human interaction benchmarks.

\end{abstract}

\input{sec/intro.tex}

\input{sec/related.tex}

\input{sec/prelim.tex}

\input{sec/method.tex}

\input{sec/results.tex}

\input{sec/conclusion.tex}

\begin{ack}
We thank David Luebke, Slim Essid, Nikhil Srihari, and Viet Anh Trinh for early discussions on the project. 
\end{ack}

{
\small
\bibliographystyle{plainnat}
\bibliography{main}
}

\appendix

\input{sec/appendix.tex}

\end{document}

%% file: macros.tex
\usepackage{amsmath}         %
\usepackage{amssymb}         %
\usepackage{mathtools}       %
\usepackage{bm}              %
\usepackage{graphicx}        %
\usepackage{subcaption}      %
\usepackage{multirow}        %
\usepackage{float}           %
\usepackage{siunitx}         %
\usepackage{xspace}          %
\usepackage[capitalise,noabbrev]{cleveref} %
\usepackage{wrapfig}         %

\graphicspath{{fig/}{figures/}{figs/}}

\newcommand{\R}{\mathbb{R}}

\crefname{section}{Sec.}{Secs.}
\Crefname{section}{Section}{Sections}
\crefname{subsection}{Sec.}{Secs.}
\Crefname{subsection}{Section}{Sections}
\crefname{figure}{Fig.}{Figs.}
\Crefname{figure}{Figure}{Figures}
\crefname{table}{Tab.}{Tabs.}
\Crefname{table}{Table}{Tables}
\crefname{equation}{Eq.}{Eqs.}
\Crefname{equation}{Equation}{Equations}
\crefname{appendix}{App.}{Apps.}
\Crefname{appendix}{Appendix}{Appendices}

\newif\ifdraft
\drafttrue

\ifdraft
  \newcommand{\todo}[1]{{\color{red}\textbf{[TODO:}\,#1\textbf{]}}}
  \newcommand{\fixme}[1]{{\color{red}\textbf{[FIXME:}\,#1\textbf{]}}}
  \newcommand{\note}[1]{{\color{blue}\textbf{[note:}\,#1\textbf{]}}}
  \newcommand{\kn}[1]{{\color{orange}\textbf{[kn:}\,#1\textbf{]}}}
  
  \newcommand{\tianye}[1]{{\color{violet}\textbf{[tianye:}\,#1\textbf{]}}}
\else
  \newcommand{\todo}[1]{}
  \newcommand{\fixme}[1]{}
  \newcommand{\note}[1]{}
  \newcommand{\kn}[1]{}
  \newcommand{\tianye}[1]{}
\fi

%% file: sec/intro.tex
\section{Introduction}

In natural dyadic interaction, listening and expressing are never discrete alternating turns, but rather a highly synchronized, streaming process encompassing both speech and physical movements. Although recent \textit{full-duplex speech} models \cite{defossez2024moshi, roy2026personaplexvoicerolecontrol} have facilitated fluid verbal exchanges, genuine interaction is fundamentally multi-modal. Establishing true social presence demands the tight coupling of speech and motion. This \textit{full-duplex speech-and-motion} mutual perception—where participants simultaneously react to verbal and physical cues—is essential for capturing the intricate dynamics of spontaneous encounters. These capabilities are foundational for next-generation Embodied Conversational Agents (ECAs) to navigate complex social nuances, directly enabling responsive human-robot interaction and high-fidelity synthetic dyadic data generation (e.g., Fig.~\ref{fig:teaser}).

Recently, several methods have advanced dyadic interaction by incorporating both speech and motion (see Table \ref{tab:capabilities}). However, they still fall short of true \textit{full-duplex speech-and-motion}, which requires simultaneously perceiving the partner's multi-modal cues (speech and motion) while generating the agent's reactive responses in a streaming, causal manner. Specifically, prior works such as Audio2Photoreal (A2P) \cite{ng2024audio2photoreal} and DyaDiT \cite{peng2026dyadit} rely on non-causal diffusion models; while achieving high visual quality, they are restricted to offline generation and are fundamentally precluded from streaming interaction. Similarly, ViBES \cite{zhang2026vibes} employs an LLM backbone to jointly model speech and motion, but utilizes a non-causal motion tokenizer and remains entirely blind to the partner's physical movements. Conversely, concurrent causal methods like SARAH \cite{ng2026sarah} and MIBURI \cite{mughal2026miburi} support streaming generation but suffer from severe perception deficits. SARAH perceives only limited spatial signals—such as the partner's 2D floor position—ignoring the semantic richness of actual body gestures. MIBURI successfully achieves full-duplex interaction in the \textit{speech} domain, yet its motion generation remains entirely \textit{monadic} (single-person). Because MIBURI cannot receive the partner's motion as input, it generates agent actions conditioned solely on speech.
\begin{table}[H]
  \vspace{-4mm}
  \centering
  \footnotesize %
  \setlength{\tabcolsep}{3pt}  
  \renewcommand{\arraystretch}{1.15}
  \caption{\textbf{Capability comparison across representative systems for human--AI dyadic interaction.} 
    The \emph{Full-duplex} column reports the modality in which the system simultaneously \emph{perceives the user} and \emph{generates the agent} in a streaming, causal manner. 
    \textbf{DyaPlex} is the only method whose motion pathway is full-duplex (perceives user motion \emph{and} streams agent motion), in addition to inheriting full-duplex speech from its backbone. Furthermore, it is trained on a corpus roughly $20$--$500\times$ larger than the corpora used by prior work. $\dagger$ SARAH perceives user's 2D floor position but not gestures.}
  \label{tab:capabilities}
  \begin{tabular}{@{} l cccc ccc l @{}} 
    \toprule
    & \multicolumn{2}{c}{Perception} & \multicolumn{2}{c}{Generation} & & & & \\
    \cmidrule(lr){2-3} \cmidrule(lr){4-5}
    Method
      & \makecell{Partner\\Speech}
      & \makecell{Partner\\Motion}
      & \makecell{Agent\\Speech}
      & \makecell{Agent\\Motion}
      & Dyadic
      & Causal
      & Full-duplex
      & \makecell{Training\\Data} \\
    \midrule
    SARAH~\cite{ng2026sarah}           & Yes & No$^\dagger$ & No  & Yes & Yes & Yes & No & Embody 3D (50\,h) \\
    A2P~\cite{ng2024audio2photoreal}   & Yes & No           & No  & Yes & Yes & No  & No & Custom (8\,h) \\
    DualTalk~\cite{peng2025dualtalk}   & Yes & Yes          & No  & Yes & Yes & No  & No & DualTalk (50\,h) \\
    ViBES~\cite{zhang2026vibes}        & Yes & No           & Yes & Yes & No  & No  & No & Converse 3D (1{,}000\,h) \\
    MIBURI~\cite{mughal2026miburi}     & Yes & No           & Yes & Yes & No  & Yes & Speech & BEAT2 (70\,h) \\
    DyaDiT~\cite{peng2026dyadit}       & Yes & Yes          & No  & Yes & Yes & No  & No & Seamless (182\,h) \\
    \midrule
    \rowcolor{gray!15} \textbf{DyaPlex (Ours)} & \textbf{Yes} & \textbf{Yes} & \textbf{Yes} & \textbf{Yes} & \textbf{Yes} & \textbf{Yes} & \textbf{Speech \& Motion} & \textbf{Seamless (4{,}000\,h)} \\
    \bottomrule
  \end{tabular}
  \vspace{-4mm}
\end{table}

As highlighted in Table \ref{tab:capabilities}, the absence of a unified, causal perception-generation loop across both modalities fundamentally breaks the natural communication flow. Consequently, systems without partner motion perception struggle to produce reciprocal behaviors jointly driven by speech and motion. To illustrate this critical bottleneck, consider conversational dynamics like \textit{subconscious mirroring} and \textit{silent backchanneling}. In natural encounters, a listener often continuously copies a speaker's posture or provides sustained motion feedback (e.g., vigorous nodding) \textit{while} actively listening, without interrupting the speech flow. Simultaneously, the speaker perceives these motion responses and dynamically adjusts their own behavior, creating a truly natural, closed-loop interaction. Traditional frameworks—forced to act as delayed turn-takers or conditioned solely on speech—remain completely oblivious to these concurrent motion cues. In such scenarios, methods failing to simultaneously perceive both the partner's speech and motion break down entirely.

To bridge this gap, we present \textbf{DyaPlex}, the first streaming, full-duplex speech-and-motion model for dyadic interaction. By seamlessly conditioning the agent's actions on concurrent multi-modal cues (speech and motion) at the frame level, it enables deeply coupled social behaviors unsupported by prior models. Our method employs a dual-tower Transformer architecture that adds motion modeling capabilities to a full-duplex speech backbone (e.g., PersonaPlex \cite{roy2026personaplexvoicerolecontrol}) while preserving its conversational reasoning. Specifically, we utilize the frozen speech model as the \textit{speech tower}. This tower processes dyadic audio and extracts rich, per-layer residual-stream hidden features that capture the linguistic and prosodic nuances of the conversation. These features are then injected into a trainable \textit{motion tower} via cross-attention.  Crucially, the motion tower embeds both participants' physical motions within a unified, autoregressive sequence. This design enables the motion features of the partner and the agent to interact deeply through self-attention, while the cross-attention mechanism ensures these representations attend to their corresponding speech signals. This dual-attention synergy not only aligns the generated motion with the speech but also fully exploits the rich prior knowledge of the foundational speech model (see Fig.~\ref{fig:teaser}). Moreover, to ensure precise temporal alignment between the speech and motion streams, we introduce cross-attention with  time-aligned RoPE~\cite{rope2024}. By injecting explicit relative temporal distances, this design effectively synchronizes the speech-motion modalities and prevents the cross-modal mapping from degenerating into time-agnostic, fixed speech feature retrieval. Finally, because the motion tower maintains strict causality—consistent with the foundational speech backbone—DyaPlex inherently supports seamless  streaming generation.

Trained on the 4,000-hour Seamless Interaction~\cite{seamless2025} dataset--a scale roughly 20 to 500 times larger than previous dyadic motion corpora--DyaPlex captures the vast diversity and high-order dependencies inherent in natural human coordination. Extensive evaluations demonstrate that our model establishes new state-of-the-art performance across both monadic and dyadic human interaction benchmarks, yielding unprecedented realism, synchrony, and socially coherent behaviors.

The primary contributions of this work are summarized as follows:
\begin{itemize}
    \item We introduce DyaPlex, the first full-duplex model capable of streaming, simultaneous perception and generation of both speech and motion for human-agent interaction.
    
    \item We propose a novel dual-tower architecture that deeply couples a trainable motion pathway with a frozen full-duplex speech model. This is achieved via unified dyadic token interleaving and a time-aligned speech-motion RoPE to ensure precise cross-modal synchronization.
    \item Extensive evaluations show our model establishes new state-of-the-art results on both monadic and dyadic benchmarks, unlocking novel applications such as responsive social robotics and scalable synthetic data generation.
\end{itemize}

%% file: sec/related.tex
\section{Related Work}

\subsection{Single-Person and Co-Speech Motion Generation}
Early 3D motion synthesis focused on single-person generation from text~\cite{lucas2022posegpt, tevet2022motionclip, jiang2023motiongpt, zhang2023generating} or speech~\cite{habibie2021learning, liu2022beat, yi2023talkshow, zhu2023taming}. While recent models~\cite{liu2024emage, mughal2025retrieving, liu2025gesturelsm} significantly advance fine-grained audio-motion synchronization, they remain strictly monadic. Concurrent attempts to unify multi-modal inputs, such as ViBES~\cite{zhang2026vibes}, fuse speech and motion inside a single backbone. However, ViBES is non-causal for tokenizer and hence non-streaming, and forcing all modalities into a shared backbone via self-attention severely limits the motion context window.  

\subsection{Human Interaction and Reaction Generation}
Beyond single characters, prior works forecast multi-person trajectories~\cite{joo2017panoptic}, synthesize interactions from text~\cite{wang2024intercontrol, liang2024intergen}, or generate physically plausible action-reaction flows (e.g., dodging, handshaking) conditioned on a partner's motion~\cite{cen2025ready_to_react, Jiang2025ARFlowHA, xu2024regennet, ghosh2024remos}. While excelling at spatial coordination, these methods operate in silent, non-conversational environments. Focusing strictly on pure motion-to-motion kinematics, they struggle to generalize to conversational dyadic settings where non-verbal social nuances and continuous spoken dialogue are deeply intertwined.

\subsection{Dyadic Conversational Human Interaction}
The most relevant domain to our work is dyadic conversational interaction, requiring the simultaneous coordination of verbal and physical behaviors. Early models focused on localized listener feedback~\cite{ng2022learning2listen}. Subsequent full-body dyadic models prioritize synthesis quality over real-time interactivity, relying on offline audio processing~\cite{ng2024audio2photoreal}, bidirectional recurrent networks~\cite{peng2025dualtalk}, or non-causal diffusion frameworks (e.g., DyaDiT~\cite{peng2026dyadit}), which fundamentally precludes streaming applications.

Recent concurrent work proposed causal methods to achieve real-time performance, but exhibit critical multi-modal perception gaps. SARAH~\cite{ng2026sarah} streams dyadic motion but only perceives the user's 2D floor-projected position, entirely ignoring the semantic richness of upper-body gestures. Conversely, MIBURI~\cite{mughal2026miburi} achieves full-duplex speech interaction, yet its gesture generation remains entirely monadic; without perceiving the partner's motion, it fails to produce visually driven reciprocal behaviors. DyaPlex addresses these limitations directly. By operating on a dyadically-interleaved motion stream within a novel dual-tower architecture, it is the first to achieve a true full-duplex loop—simultaneously perceiving partner full-body motion and speech and streaming multi-modal responses causally.

%% file: sec/prelim.tex
\section{Preliminaries}
\label{sec:prelim}

DyaPlex couples two components: an off-the-shelf full-duplex speech model (PersonaPlex~\cite{roy2026personaplexvoicerolecontrol}) that remains frozen, and a body-part-aware streaming RVQ-VAE motion tokenizer that we pre-train independently. We summarize their key features below.

\subsection{PersonaPlex: Full-Duplex Speech Tower} 
\label{sec:prelim_pp}

We use PersonaPlex~\citep{roy2026personaplexvoicerolecontrol}, a causal Transformer built on Moshi~\citep{defossez2024moshi}, which is a full-duplex speech-language architecture featuring a hidden dimension of $d_s=4096$ and $L_s=32$ layers (see Fig.~\ref{fig:arch} (b)). Both speakers' speech is tokenized at $f_s{=}12.5$\,Hz with the Mimi
neural speech codec~\citep{defossez2024moshi}. At frame $t$, PersonaPlex receives a $17$-way interleaving of text and dyadic speech codebooks. At the embedding layer, these $17$ per-codebook embeddings are summed into a single $d_s$-dim vector per frame. We freeze PersonaPlex throughout our training. Crucially, for every transformer block $\ell{=}1,\dots,L_s$, PersonaPlex exposes its post-block residual-stream hidden states $\mathcal{H}_\ell \in \mathbb{R}^{T \times d_s}$, where $T$ is the sequence length. These hierarchical hidden states capture the rich conversational context and are directly consumed by our motion tower (\cref{sec:method}).

\subsection{Body-part-based RVQ-VAE Motion Tokenizer}
\label{sec:prelim_rvq}

\begin{figure*}[t]
  \centering
  \includegraphics[width=0.95\linewidth]{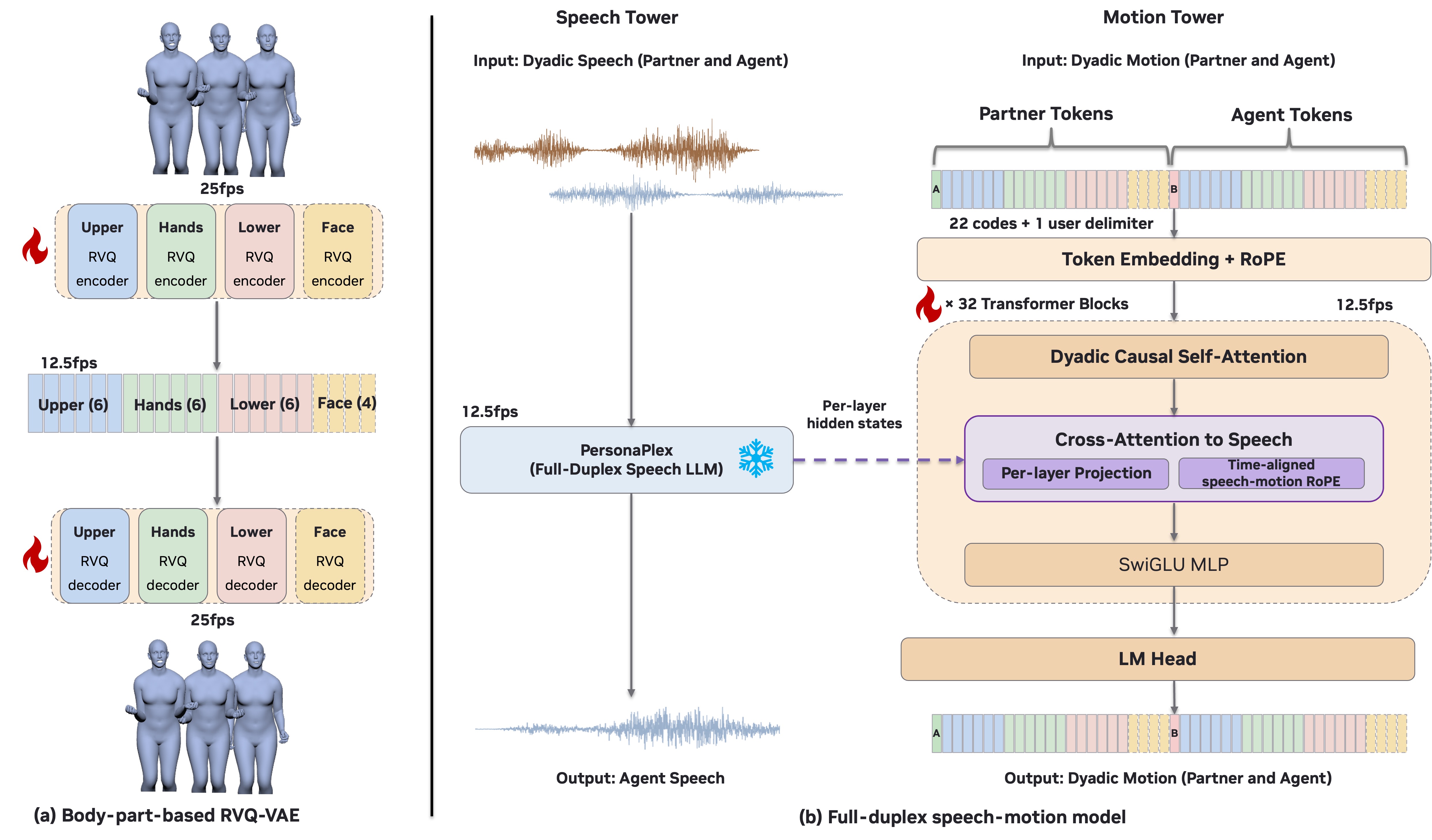}
  \caption{\textbf{Architecture overview.}
    DyaPlex consists of three components: (a)  part-aware RVQ-VAE decoders and (b) a frozen speech tower, and a trainable motion tower. The speech tower (PersonaPlex) takes in dyadic speech, emits agent speech autoregressively, and exposes its per-layer residual-stream hidden states $\{\mathcal{H}_\ell\}_{\ell=1}^{32}$. For training, we precompute $\{\mathcal{H}_\ell\}$ once (\cref{sec:hidden_extraction}) to serve as cross-attention keys and values for the motion tower.  
    The $32$-layer causal motion tower ($d_m{=}1024$, $h_m{=}16$) operates on a dyadically interleaved stream at $12.5$\,fps: $[\mathtt{[A]}, \mathbf{m}^A_t, \mathtt{[B]}, \mathbf{m}^B_t, \dots]$. Each block applies dyadic causal self-attention on the motion stream, followed by cross-attention to the speech states using a learned projection ($h_c{=}12$, $d_c{=}64$) and \emph{time-aligned speech-motion RoPE}. The LM head outputs motion tokens (supervised on 18 SMPL-H body codes for our body-only base model) at $12.5$\,fps, which the RVQ-VAE decoders with $2{\times}$ temporal upsampling then finally reconstruct into SMPL-X pose parameters at 25\,fps.
  }
  \label{fig:arch}
  \vspace{-6mm}
\end{figure*}

We tokenize body and (optionally) face motion with a body-part-aware RVQ-VAE, adapted from GestureLSM~\citep{liu2025gesturelsm}. We retrain the body tokenizer end-to-end on the Seamless Interaction dataset~\citep{seamless2025} and modify it into a causal streaming architecture: the encoder consumes 25\,fps motion as input and outputs tokens at the speech-aligned rate $f_m{=}12.5$\,fps, while the decoder applies $2{\times}$ temporal upsampling to reconstruct 25\,fps SMPL-X output. Our RVQ-VAE consists of four independent decoders specialized to the upper body, hands, lower body, and face (see Fig.~\ref{fig:arch} (a)). Each frame is encoded by $K{=}22$ codes ($18$ body codes and $4$ face codes) drawn from a shared vocabulary of size $V_\text{mot}{=}4096$, which is partitioned into four disjoint $1024$-entry bands. At inference, these codes are routed to their respective decoders to reconstruct SMPL-X~\cite{SMPL-X:2019} and FLAME~\cite{FLAME:SiggraphAsia2017} parameters. Both the encoder and decoder are frozen during the subsequent motion-tower training in Fig.~\ref{fig:arch} (b). For complete details regarding the adaptation of the causal architecture, the end-to-end streaming training procedure, and the decoding of tokens into 3D meshes, please refer to the supplement.

%% file: sec/method.tex
\section{Full-duplex speech-motion model}
\label{sec:method}

We frame full-duplex dyadic interaction as time-aligned autoregressive speech and motion generation for both interacting partners via dedicated speech and motion towers. A frozen speech tower (\cref{sec:prelim_pp}) models
audio; a trainable motion tower cross-attends to the speech tower's
per-layer hidden features through learned key/value projections. Our motion tower
generates motion for \emph{both} speakers of the dyad in a single
interleaved stream, enabling the agent to both simultaneously \emph{perceive} the
partner's body motion and \emph{respond} through its own.
\cref{fig:arch} summarizes the design.

\paragraph{Notation.}
We consider a dyad $(A, B)$ where, without loss of generality, $A$ is
the partner and $B$ the agent. Their Mimi audio tokens are
$\mathbf{s}^A_{1:T} , \mathbf{s}^B_{1:T}$ at $f_s{=}12.5$\,Hz; their
RVQ-VAE motion tokens are $\mathbf{m}^A_{1:T}, \mathbf{m}^B_{1:T}$
at $f_m{=}12.5$\,fps, $T$ is the total number of frames. Each motion frame is a $K{=}22$-dim integer vector
$\mathbf{m}_t = (c^{(1)}_t, \dots, c^{(K)}_t)$ over a shared vocabulary
of size $V_\text{mot}{=}4096$.

\subsection{Speech hidden-state extraction with a hybrid input prefix}
\label{sec:hidden_extraction}

The motion tower conditions on PersonaPlex's per-layer residual-stream
hidden states $\{\mathcal{H}_\ell\}_{\ell=1}^{L_s}$ via cross-attention
at every block. These hidden states encode
the joint conversational context --- both speakers' acoustic content,
PersonaPlex's inner-monologue text predictions, and the implicit
turn-taking dynamics of the dyadic input format. Exposing all
$L_s{=}32$ layers rather than a single final embedding gives the motion
tower's cross-attention access to all intermediate speech
representations, not only the final-layer embedding. We pair
motion-tower blocks one-to-one with PersonaPlex layers
(\cref{sec:motion_tower}), letting cross-attention learn its own hierarchical speech representations.
For training, the causal architecture of PersonaPlex allows us to precompute all hidden states in a single teacher-forced forward pass. Conversely, during inference, these states are generated autoregressively on the fly using the standard PersonaPlex inference loop.
\paragraph{Hybrid system prompt alignment.}
PersonaPlex requires a hybrid system prompt (text and voice) to initialize its auto-regressive generation. To match this distribution of the pre-trained PersonaPlex at training time, we explicitly prepend a constructed system prompt to each Seamless clip before extracting the hidden states. Please refer to the supplement for more details. 

\subsection{Full-duplex motion tower}
\label{sec:motion_tower}

The motion tower is a causal decoder-only Transformer with
$L_m{=}32$ blocks (one per PersonaPlex layer), dimension $d_m{=}1024$,
$h_m{=}16$ self-attention heads of head dimension $d_m / h_m {=} 64$,
RoPE~\cite{rope2024} self-attention, and SwiGLU~\cite{shazeer2020gluvariantsimprovetransformer} feed-forward layers. A single shared
embedding $\mathbf{E} \in \R^{V \times d_m}$ maps token ids to features, where the vocabulary
$V = V_\text{mot} + 2$ includes two special speaker tags
$\mathtt{[A]}, \mathtt{[B]}$ (see Fig.~\ref{fig:arch}).

\paragraph{Self-attention with dyadic interleaving.}

We flatten the two motion streams into a single token sequence that
alternates speaker tags and their $K$ RVQ codes at every frame:
\begin{equation}
\label{eq:interleave}
\mathbf{M} = \big[\,
  \underbrace{\mathtt{[A]},\, \mathbf{m}^A_1,\, \mathtt{[B]},\, \mathbf{m}^B_1}_{\text{frame 0}},\;
  \underbrace{\mathtt{[A]},\, \mathbf{m}^A_2,\, \mathtt{[B]},\, \mathbf{m}^B_2}_{\text{frame 1}},\;\dots\,
\big].
\end{equation}
This arrangement yields a per-frame ``step'' length of $L_\text{step} = 2(K + 1) = 46$ tokens. 
Consequently, applying causal self-attention over the unified sequence $\mathbf{M}$ simultaneously models three distinct dependencies: (i) the intra-frame coherence between a single speaker's $K$ RVQ codes, (ii) the within-frame cross-speaker reactions ($A \to B$), and (iii) the long-range temporal dynamics across consecutive frames. Ours is the first
architecture in which both sides of a dyadic conversation share a single
autoregressive motion prior. Adding partner motion information dramatically improves the dyadic motion quality compared to our model without partner as measured in P-FD and $\Delta$-User experiment in Tab~\ref{tab:t2_dyadic_fullduplex}. 

\paragraph{Cross-attention with time-aligned speech-motion RoPE.}
Within the motion tower, each transformer block $\ell$ is paired one-to-one with a corresponding PersonaPlex block. These towers interact through a multi-head cross-attention sub-layer with $h_c{=}12$ heads of per-head dimension $d_c{=}64$, where queries are derived from the interleaved motion stream, and keys/values are computed from the frozen speech hidden states $\mathcal{H}_\ell$. Specifically, at each block $\ell$, the motion-tower hidden state $\mathbf{h}_t \in \mathbb{R}^{d_m}$ at flattened token position $t$ in $\mathbf{M}$ is projected by a trainable matrix $\mathbf{W}_q^\ell \in \mathbb{R}^{d_m \times h_c d_c}$ to produce the query $\mathbf{q}_t$. Similarly, trainable matrices $\mathbf{W}_k^\ell, \mathbf{W}_v^\ell \in \mathbb{R}^{d_s \times h_c d_c}$ project $\mathcal{H}_\ell$ into keys $\mathbf{K}_\ell$ and values $\mathbf{V}_\ell$. This learned-projection architecture provides the motion tower with the capacity to remap speech features from the pre-trained PersonaPlex---which were natively optimized for audio synthesis---into representations tailored for gesture generation. 

Crucially, to temporally align the motion and speech modalities, we assign every motion token a query position $q_{\text{pos}}(t)$ corresponding to its actual frame index. Specifically:
\begin{equation}
\label{eq:qpos}
q_{\text{pos}}(t) \;=\; \big\lfloor t \,/\, L_\text{step} \big\rfloor.
\end{equation}
Consequently, all $L_\text{step}{=}46$ tokens comprising a single motion frame share the exact same query position (e.g., all tokens in the $0$-th frame of $\mathbf{M}$ are assigned $q_{\text{pos}} = 0$). We then apply Rotary Positional Embeddings (RoPE~\cite{rope2024}) to the queries and keys using these positions:
\begin{equation}
\label{eq:rope}
\begin{aligned}
\tilde{\mathbf{q}}_t &= \mathrm{RoPE}\!\big(\mathbf{q}_t,\, q_{\text{pos}}(t)\big), \\
\tilde{\mathbf{k}}_s &= \mathrm{RoPE}\!\big(\mathbf{K}_\ell[s],\, s\big).
\end{aligned}
\end{equation}
where $s$ denotes the temporal index of the speech tokens. Since the motion sampling rate matches the speech sampling rate ($f_m = f_s$), the indices $q_{\text{pos}}(t)$ and $s$ operate on a \textit{unified} temporal axis.
For each block $\ell$, we define $\tilde{\mathbf{K}}_\ell = [\tilde{\mathbf{k}}_1, \dots, \tilde{\mathbf{k}}_T]$ as the full sequence of rotated keys obtained by applying RoPE to each projected speech state. The cross-attention output for the $\ell$-th layer is then calculated as $\mathrm{XAttn}^\ell(\mathbf{h}_t) = \mathrm{Attention}\!\big(\tilde{\mathbf{q}}_t,\, \tilde{\mathbf{K}}_\ell,\, \mathbf{V}_\ell\big)$. Since the RoPE computes attention based strictly on the relative offset $q_{\text{pos}}(t) - s$, the ideal solution simply reduces to a diagonal alignment, where each motion frame attends directly to its concurrent speech frame. This provides an inductive bias toward time-aligned attention, which the network learns to exploit during training. We refer to this mechanism as \emph{time-aligned speech-motion RoPE}. Without RoPE, the cross-attention has no explicit positional signal and would have to recover the alignment implicitly through the motion query and speech key alone, which results in suboptimal speech-motion alignment. The importance of both cross attention and this speech-motion alignment via RoPE is empirically validated by the clear degradation in the BeatAlign score when they are removed (see Tab.~\ref{tab:t2_dyadic_fullduplex}; further analysis is provided in the supplement).

\paragraph{Speech context window and causality.}
To ensure strict causality for real-time streaming, the cross-attention mechanism dictates that a motion token $t$ can only attend to speech frames at or preceding its concurrent motion frame. This constraint is enforced via a causal mask $M_{t,s}$:
\begin{equation}
\label{eq:causal_mask}
M_{t,s} \,=\, 
\begin{cases} 
1, & \text{if } s \le q_{\text{pos}}(t) \\ 
0, & \text{otherwise} 
\end{cases}
\end{equation}
Theoretically, cross-attention can access an indefinite speech history without added motion self-attention overhead. For simplicity, we align the speech context with the motion tower's $4096$-token window, bounding the receptive field to $89$ frames ($\approx 7.1$\,s at $12.5$\,fps).

\subsection{Training Objective}\label{sec:training}We optimize the motion tower parameters $\theta$ via teacher-forced next-token prediction over the interleaved sequence $\mathbf{x}$, conditioned on the precomputed speech states $\{\mathcal{H}_\ell\}$. To prevent the model from assigning probability mass to structurally invalid tokens across the four disjoint RVQ codebooks, we apply a band-mask (setting out-of-band logits to $-\infty$) prior to the softmax. The masked cross-entropy loss is:$$\mathcal{L}_\text{CE}(\theta)=- \sum_{t \in \mathcal{S}} \log p_\theta( x_{t+1} | x_{\le t}; \{\mathcal{H}_\ell\}),$$where $\mathcal{S}$ denotes the $18$ supervised body code positions per frame using the official SMPL-H body data (body + hands) that comes with the Seamless~\citep{seamless2025} dataset. In this paper we focus on body motion generation; throughout, ``Ours" refers to this \emph{body-only} base model. A separate variant of our model that also predicts face codes is used only for the qualitative demos in Fig.~\ref{fig:teaser} and the supplementary video. Additionally, following MIBURI~\citep{mughal2026miburi}, we employ a linear voice-activation head to predict the binary speaking/listening state $v_t$ at each valid code position $t \in \mathcal{V}$. This yields an auxiliary binary cross-entropy loss $\mathcal{L}_\text{VA}$. Our final training objective is simply $\mathcal{L}=\mathcal{L}_\text{CE} + \beta \mathcal{L}_\text{VA}$, where we empirically set $\beta=0.01$.

\subsection{Streaming inference}
\label{sec:inference}

During inference, the model operates as a causal streaming sampler.
The PersonaPlex speech tower synthesizes agent B's speech while also generating the per-layer hidden states $\mathcal{H}_\ell$ that encode the dyadic conversational context.
Conditioned on $\mathcal{H}_\ell$, the motion tower can either generate both speakers' motion jointly or only the agent's motion, depending on the application.
In the \emph{both-speaker mode} (e.g., synthetic dyadic interaction data generation), the motion tower autoregressively samples motion tokens at both $\mathtt{[A]}$ and $\mathtt{[B]}$ slots, producing complete dyadic motion for both speakers.
In the \emph{agent-only mode} (e.g., human-agent / robot interaction), given an observed partner-motion prefix $\mathbf{m}^A_{1:f}$ and the PersonaPlex hidden states $\mathcal{H}_\ell$ for both speakers up to frame $f$, we fill in the observed partner tokens into the $\mathtt{[A]}$ slots and sample from the learned distribution $p_\theta$ exclusively at the agent's $\mathtt{[B]}$ slots:
\begin{equation}
\label{eq:agent_sampling}
\hat{c}^{(k)}_f \,\sim\, \mathrm{topk}\!\big(\mathrm{softmax}(\mathrm{logits}(\mathbf{x}_{<t}) / \tau),\; K_\text{top}\big),
\end{equation}
where $\mathbf{x}_{<t}$ denotes the partial interleaved sequence up to flat position $t$ (\cref{eq:interleave}). To rigorously maintain the dyadic structure, we deterministically insert the appropriate speaker tags ($\mathtt{[A]}$ or $\mathtt{[B]}$) and copy the ground-truth partner motion $\mathbf{m}^A_f$ at their designated positions. The autoregressive sampling is restricted entirely to the $\mathtt{[B]}$ code positions, applying a temperature of $\tau=1.0$ and $K_\text{top}=200$ throughout.

Crucially, because the speech hidden states $\mathcal{H}_\ell$ are continuously produced by a frozen streaming speech tower, and the part-aware RVQ decoders (\cref{sec:prelim_rvq}) are inherently causal, the entire generation pipeline---from partner audio input, through PersonaPlex and the motion tower, down to the final SMPL-X reconstruction---maintains strict causality. This architectural guarantee allows the system to be executed chunk-wise, enabling genuine real-time streaming interaction.

%% file: sec/results.tex
\section{Experiments}

\paragraph{Datasets.}
We utilize the Seamless Interaction dataset~\cite{seamless2025}, a large-scale corpus featuring approximately \SI{4000}{\hour} of dyadic conversations. After applying filters to remove invalid and corrupted data, \num{57947} pairs (\SI{3435}{\hour}) of dyadic motion are available for training. For evaluation, we further restrict to a \num{330}-pair test subset (\SI{\sim 18}{\hour}), retaining only pairs where both speakers pass additional audio quality check (e.g., broken or missing recordings). We used the first 20 seconds of the test set for the evaluations. Both audio and motion are tokenized at a synchronized rate of \SI{12.5}{\hertz} using Mimi~\cite{defossez2024moshi} and our causal RVQ-VAE, respectively. For  details on data processing pipeline, please refer to the supplement.
\paragraph{Baselines.}
We compare our method with Audio2Photoreal~\cite{ng2024audio2photoreal} (A2P) and DualTalk~\cite{peng2025dualtalk}, state-of-the-art dyadic conversational models using official source code. A2P is representative for a diffusion-based method and DualTalk for a transformer-based method. We adapt both baselines and retrain all methods, including our body-only base model, on the same Seamless SMPL body dataset (see the supplement for more details).

\paragraph{Metrics.}
We report two groups of metrics on the Seamless test set, summarized in \cref{tab:t2_dyadic_fullduplex}: (i)~\emph{Monadic \& Alignment Metrics}: FGD~\cite{yoon2020trimodal} and Diversity~\cite{ng2024audio2photoreal} evaluate individual motion quality, while BeatAlign~\citep{aichoreo,liu2022beat} measures speech-motion synchronization. (ii)~\emph{Dyadic Interaction Metrics}: Paired Fr\'echet Distance (P-FD)~\citep{ng2022learning2listen, maluleke2024synergy} evaluates the realism of the joint user--agent motion distribution. User-Conditioning Gain ($\Delta$-User) quantifies the model's reliance on partner motion, defined as the relative P-FD improvement when conditioning on matched versus shuffled user motions: $\Delta_\text{User} = (\mathrm{P\text{-}FD}_\text{shuffled} - \mathrm{P\text{-}FD}_\text{matched}) / \mathrm{P\text{-}FD}_\text{matched}$ expressed as a percentage. By definition, baselines lacking a user-motion pathway inherently score $0\%$. Since A2P and DualTalk uses ground-truth audio of partner and agents as input, for a fair comparison, we evaluate our model under a \emph{teacher-forced} configuration: ground-truth PersonaPlex hidden states and user motion tokens are provided, and only the agent's motion tokens are sampled. All evaluations are conducted at \SI{25}{fps}. Ground-truth body joints are derived from the Seamless SMPL-H parameters via SMPL-X forward kinematics. For evaluation, following Seamless~\cite{seamless2025}, we set the root translation to zero for all methods. All metrics evaluate body motion and interaction quality using the 22 SMPL-X body joints (excluding 
  finger and face joints). 

\begin{figure}[t]
  \centering
  \includegraphics[width=0.9\linewidth]{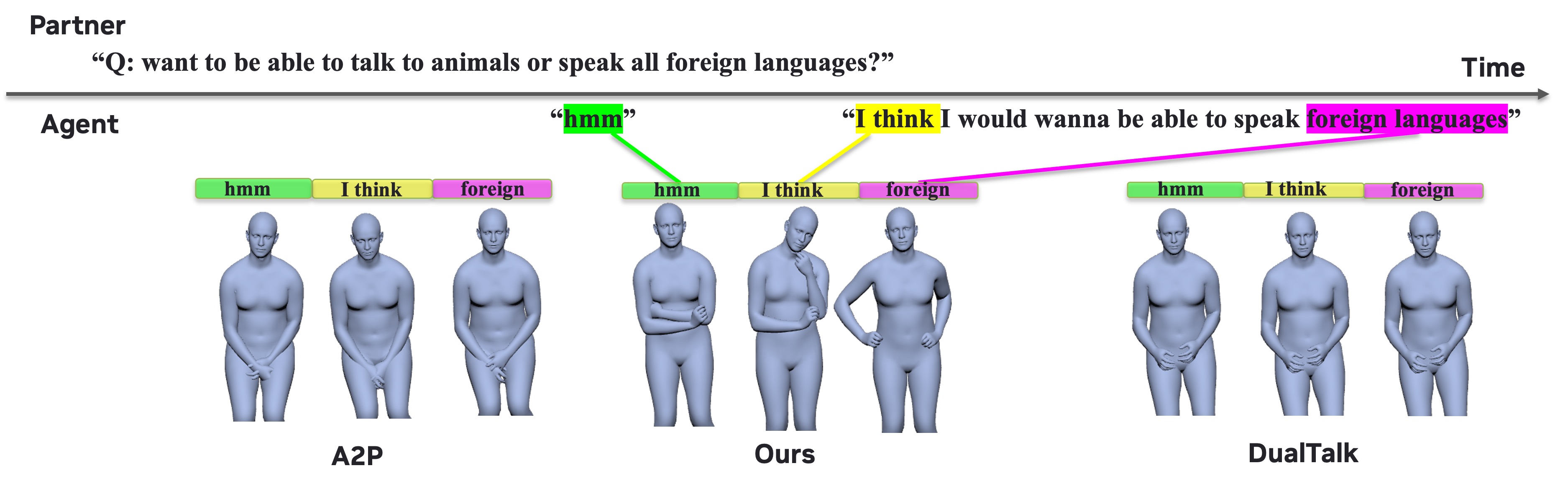}
  \vspace{-4mm}
  \caption{Qualitative comparison on Seamless test clips.
  The agent (speaker~$B$) is generated by a different method, conditioned on the same ground-truth user (speaker~$A$) motion and speech.
  Compared with Audio2Photoreal~\citep{ng2024audio2photoreal} and DualTalk~\cite{peng2025dualtalk} (frozen due to mode collapse), our model produces more diverse and natural dyadic behaviors.}
  \label{fig:qual_comparison}
  \vspace{-4mm}
\end{figure}

\begin{figure}[t]
  \centering
  \includegraphics[width=0.9\linewidth]{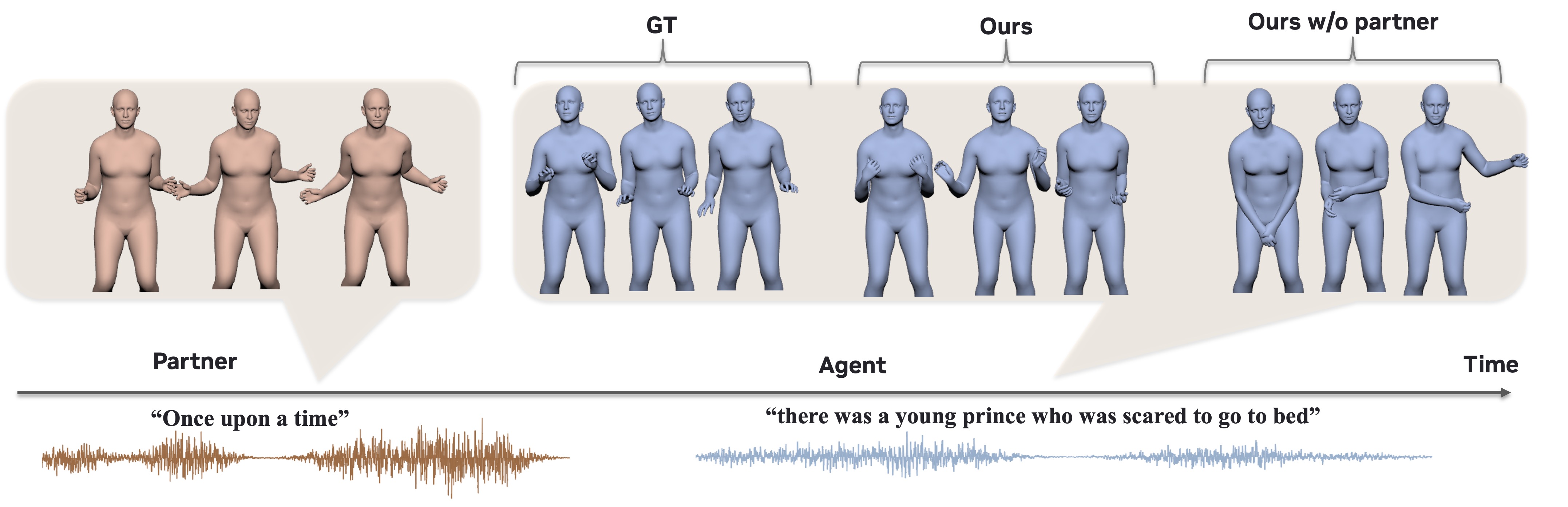}
  \vspace{-4mm}
  \caption{In this example the pair reads a story in a turn taking fashion and they use body languages to communicate turn taking. Our dyadic model produces more coherent gestures with the partner and the ground truth agent. Our model without perceiving the partner motion (Ours w/o partner) fails to produce coherent gestures as it cannot see the partner gestures.  }
  \label{fig:ablation}
  \vspace{-4mm}
\end{figure}

\begin{table}[t]
\centering
\small %
\setlength{\tabcolsep}{4pt} %
\renewcommand{\arraystretch}{1.15} %
\vspace{-4mm}
\caption{\textbf{Quantitative results on Seamless.}%
~Two column groups: Monadic metrics (left) and dyadic metrics (right).
Columns marked ``$\downarrow$'' are lower-is-better (FGD, P-FD); all other columns report values for which closer to GT is better.
$\Delta$-User cells with positive percentages indicate methods whose generated motion changes when user motion is shuffled at inference. $\dagger$ indicates methods do not use partner motion information as input and thus $0\%$ by design. \textbf{Bold} and \underline{underline} indicate best and second-best per column among baselines and Ours variants. DualTalk reports substantially worse results due to training collapse that results in frozen motion, and their BeatAlign is undefined due to no detectable motion.}
\label{tab:t2_dyadic_fullduplex}
\begin{tabular}{@{} l ccccc @{}} %
\toprule
 & \multicolumn{3}{c}{Monadic} & \multicolumn{2}{c}{Dyadic} \\
\cmidrule(lr){2-4} \cmidrule(l){5-6}
Method & \makecell{FGD~$\downarrow$ \\ \scriptsize $(\times 10^{-3})$} & \makecell{Diversity \\ \scriptsize $\rightarrow$GT} & \makecell{BeatAlign \\ \scriptsize $\rightarrow$GT} & \makecell{P-FD~$\downarrow$ \\ \scriptsize $(\times 10^{-3})$} & $\Delta$-User~$\uparrow$ \\
\midrule
GT & --- & 0.633 & 0.049 & --- & --- \\
GT (Random) & 13 & 0.683 & 0.050 & 33 & $0\%^\dagger$ \\
\midrule
\multicolumn{6}{@{}l}{\textit{Baselines}} \\
Audio2Photoreal~\citep{ng2024audio2photoreal} & 57 & 0.395 & \textbf{0.051} & 72 & $0\%^\dagger$ \\
DualTalk~\cite{peng2025dualtalk} & 161 & 0.305 & --- & 163 & +0.3\% \\
\midrule
\multicolumn{6}{@{}l}{\textit{Ablations (Ours)}} \\
\quad w/o Self-Attn & 41 & 0.416 & 0.132 & 45 & $0\%^\dagger$ \\
\quad w/o Partner & 39 & 0.725 & 0.064 & 41 & $0\%^\dagger$ \\
\quad w/o Cross-Attn & 41 & 0.708 & 0.080 & 44 & \underline{+15\%} \\
\quad w/o Cross-RoPE & \underline{8.4} & \underline{0.582} & 0.064 & \underline{10} & \textbf{+31\%} \\
\rowcolor{gray!15} \textbf{Ours (body-only)} & \textbf{5.6} & \textbf{0.611} & \underline{0.059} & \textbf{7.3} & \textbf{+31\%} \\
\bottomrule
\end{tabular}
\vspace{-4mm}
\end{table}

\subsection{Qualitative Results}
Fig.~\ref{fig:qual_comparison} shows qualitative comparisons to the baselines. Our model produces gestures that are diverse and appropriate to the conversation while baselines generate less diverse results. Fig.~\ref{fig:ablation} shows qualitative comparisons to our ablation model without partner perception (\emph{w/o Partner}). Our model without partner perception fails to mimic partner gestures. 

\subsection{Quantitative Results}
Tab.~\ref{tab:t2_dyadic_fullduplex} shows quantitative results against baselines. We additionally report ground truth (GT) and randomly sampled ground truth data (GT (Random)) as additional references. The elevated P-FD in GT (Random) confirms that P-FD works as expected to measure the joint motion distributions of two people. Our method significantly outperforms all baselines in both monadic and dyadic motion quality (FGD, P-FD). Our method has the most similar diversity to GT and scores second best BeatAlign.

\paragraph{Ablation study.}
Tab.~\ref{tab:t2_dyadic_fullduplex} also shows comparisons to our ablation models. Removing self attention (\emph
{w/o self-attention}) leads to significantly worse results across the board. In the dyadic setting, removing partner perception (\emph{w/o Partner}) degrades FGD $\sim$$7{\times}$ ($5.6 \to 39$) and P-FD $\sim$$5.6{\times}$ ($7.3 \to 41$) above the full model. Removing cross attention (\emph{w/o cross-attention}) and RoPE (\emph{w/o cross-attn RoPE}) leads to degraded BeatAlign scores due to entirely missing the speech context or imprecise alignment between speech-motion tokens. $\Delta$-User shows how much P-FD a model gains when partner motion is available. As shown in +31\% gain in P-FD in our method, in the dyadic setting, having the partner motion as input is crucial.

\paragraph{User Study.}
\begin{wraptable}{r}{0.5\textwidth} %
\centering
\footnotesize
\setlength{\tabcolsep}{10pt}
\renewcommand{\arraystretch}{1.10}
\vspace{-4mm}
\caption{
\textbf{User study of naturalness of the generated agent motion on the Seamless~\cite{seamless2025} test set}.
We report the percentage of the participants that prefer our method against the counterparts.
}
\label{tab:t6_userstudy}
\begin{tabular}{l c}
\toprule
Comparison & Preference to Ours \\
\midrule
vs. Ground Truth                                   & 29.4\% \\
vs. Audio2Photoreal~\citep{ng2024audio2photoreal}  & 66.3\% \\
vs. DualTalk~\cite{peng2025dualtalk}               & 97.5\% \\
\bottomrule
\end{tabular}
\vspace{-4mm}
\end{wraptable}
To evaluate perceived motion naturalness, we conducted a user study comparing our method against DualTalk~\cite{peng2025dualtalk} and Audio2Photoreal~\cite{ng2024audio2photoreal} on the Seamless test set~\cite{seamless2025}. Thirty-two participants used an interactive UI to compare paired mesh-rendered videos with the same ground-truth conversation audio; only the agent body motion differs between methods. As detailed in Tab.~\ref{tab:t6_userstudy}, our method is overwhelmingly preferred over DualTalk ($97.5\%$) and Audio2Photoreal ($66.3\%$), validating its superior generation quality. Notably, our generated motions achieve a $29.4\%$ preference rate even when compared directly against the ground truth.

\paragraph{Runtime performance.}

Evaluated on a single RTX A6000 Ada GPU at 12.5 Hz, the audio tower and RVQ-VAE decoder run efficiently at 30ms and 0.8ms per frame, respectively. The autoregressive motion tower is the primary computational bottleneck, taking 173ms/frame with a full 4096-token context. By reducing the motion context to 1024 tokens (1.8s) while preserving the full 7.1s speech context, the motion tower latency drops to 80ms/frame, successfully achieving real-time inference.

%% file: sec/conclusion.tex
\section{Conclusion}

We introduced DyaPlex, a full-duplex speech-and-motion framework for streaming dyadic interaction. By coupling a frozen PersonaPlex speech tower with a trainable motion tower via our time-aligned speech-motion RoPE mechanism, we achieve precise cross-modal synchronization while maintaining strict causality. This approach elegantly reduces temporal alignment to a structural inductive bias that the motion tower learns to exploit during training. Trained on the 4,000-hour Seamless Interaction dataset, our model establishes new state-of-the-art performance in both individual motion realism and joint dyadic interaction. By ensuring end-to-end causal streaming, DyaPlex enables genuine, low-latency interactions, providing a scalable and efficient foundation for next-generation embodied conversational agents.

%% file: sec/appendix.tex
\clearpage
\section{Discussion}
\label{appendix:discussion}
\subsection{Limitations and future work}
\label{appendix:limitations}
Our model currently decodes one body token at a time for 22 codes, which may not be most optimal for performance. For the actual deployment in robot, it would be worthwhile to explore chunk by chunk decoding (e.g., 6 upper body token in one go) or the decomposed depth transformer-based design similar to Moshi~\cite{defossez2024moshi} to speed up the inference. It would be also fruitful future work to explore other generative models backbone such as diffusion models. Our model can handle interactions with only 2 users. 
Extending our model to polyadic interaction would be interesting future work. 

\subsection{Potential negative societal impacts}
\label{appendix:societal}
DyaPlex outputs generic SMPL-X body pose parameters rather than photorealistic pixels, so the model itself does not directly produce identity-bearing video. However, generated motion could be paired with downstream identity-conditioned video diffusion models to produce realistic-looking videos of fabricated dyadic interactions, creating potential risks for accessible deepfakes.
We highlight a few mitigations that follow naturally from the design of DyaPlex. First, our model is \emph{identity-agnostic}: it generates motion on the generic SMPL-X skeleton without conditioning on any subject identifier, so impersonation of a specific person requires a separate identity-conditioned avatar/voice pipeline whose own safeguards apply. Additionally, there exist tools to detect videos generated by AI models based on pixel-footprints intrinsic to video generators (e.g., ~\cite{corvi2025seeing}). Second, the speech tower (PersonaPlex) is frozen and used in its public release configuration; we add no fine-tuning that targets specific real-world voices. 
On the positive side, the same capability supports beneficial applications: scalable synthetic dyadic interaction data for training social robots and embodied agents, and virtual agents and robots for full-duplex human AI interactions as discussed in the introduction.

\section{Effect of time-aligned speech-motion RoPE}
\begin{figure}
  \centering
  \includegraphics[width=\linewidth]{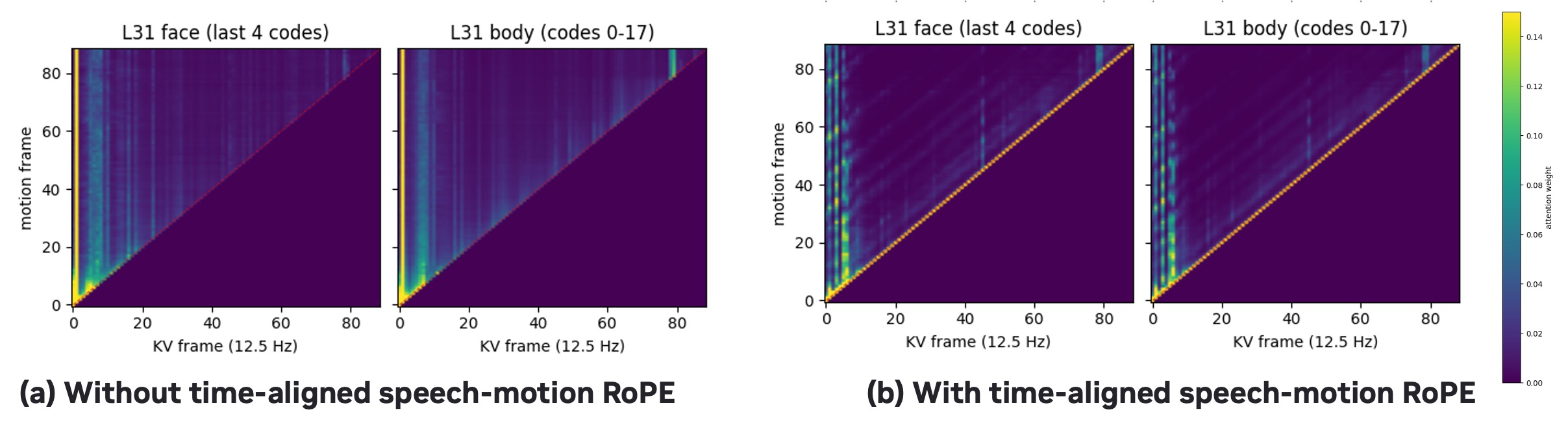}
  \caption{Comparisons without (a) and with (b) time-aligned speech-motion RoPE. }
  \label{fig:rope}
\end{figure}

Fig.~\ref{fig:rope} ablates the importance of having time-aligned speech-motion RoPE in the cross attention weight. The diagonal dotted red line shows an expected alignment. The figure shows that without RoPE the motion cannot cleanly attend to the speech features at the same time frame. With RoPE, the cross attention map lights up diagonally, demonstrating the motion tokens are correctly attending to the speech feature of that frame.

\section{Details of Seamless dataset processing}

\subsection{Dataset filtering}
\label{appendix:dataset_filtering}
Seamless Interaction~\citep{seamless2025} ships per-clip SMPL-H body
parameters recovered by per-frame
HMR2~\cite{Goel_2023_ICCV} regression and
per-frame hand parameters by HaMeR~\cite{pavlakos2024reconstructing}. Both are off-the-shelf monocular regressors;
their failure modes leak into training unless filtered explicitly.
We apply two filters: a clip-level aspect-ratio filter and a
frame-level validity mask.
\paragraph{Aspect-ratio and integrity filter (clip level).}
The Seamless videos cover a heterogeneous mix of aspect ratios.
Roughly $94.5\%$ are typical $9{:}16$ portrait clips, with smaller
fractions of square and near-square crops; the remaining ${\sim}1.5\%$
are landscape ($3840{\times}2160$, $1920{\times}1080$), rotated
($640{\times}480$), or zero-resolution unreadable / corrupt files.
HMR2 produces unusable body fits on landscape or rotated frames and
fails outright on unreadable ones. We filter at the pair level: a
dyadic pair is dropped if either speaker's video is flagged as
landscape, rotated, or unreadable, removing approximately $1.5\%$ of
the pairs in the dataset.
\paragraph{Per-frame HMR2 validity (frame level).}
Even on aspect-correct videos, HMR2 fails to detect a body in
${\sim}2.4\%$ of frames. The dataset's SMPL-H NPZs carry a per-frame
\texttt{smplh:is\_valid} flag marking these. We drop frames where
\texttt{is\_valid == 0} from all downstream processing (statistics
computation, RVQ-VAE training input, motion-tower training input,
evaluation). Pairs with fewer than $8$ valid frames after masking are
dropped entirely.

\subsection{Adding facial expressions}
\label{appendix:face_extraction}
Seamless Interaction's body annotations are SMPL-H (body $+$ hands
only) and do not include facial expression parameters. For
preliminary face experiments we augment the dataset with per-frame
FLAME~\cite{FLAME:SiggraphAsia2017} face parameters
extracted by SPECTRE~\cite{filntisis2022visual} run on the source videos. SPECTRE is a video-conditioned
monocular face reconstruction model that produces per-frame FLAME
jaw rotation, expression coefficients, and shape coefficients.
\paragraph{Face representation.}
For each Seamless video, SPECTRE outputs per-frame 6D head rotation, 6D jaw rotation, and 50D FLAME expression PCs. We retain the 6D jaw rotation and the 50D expression to form a 56-dimensional face input, discarding the 6D head rotation (redundant with the SMPL-H body's joint 0, the global root).
\paragraph{SPECTRE quality filter.}
SPECTRE produces all-zero face vectors on frames where face detection
or fitting fails. We treat any pair with ${\geq}20\%$ zero-face frames
in either speaker as unusable for face training and exclude it from
the face-codec training subset. Remaining isolated zero frames within
accepted pairs are kept as-is.
\paragraph{Scope.}
Face supervision in the Seamless dataset is preliminary. The extracted-FLAME corpus is used only for the independent RVQ-VAE encoder and decoder corresponding to the face (\cref{appendix:rvq_details}) (RVQ-VAE encoders and decoders are independent per body part) and a variant of motion tower which jointly predicts body, hands and a face and only used in the qualitative face demos in Fig.~\ref{fig:teaser} and the supplementary video.

\section{Implementation details}

\subsection{Details of PersonaPlex hybrid prompt in Seamless}
\label{appendix:hybrid_prefix}
We feed Seamless data into PersonaPlex's training-time
\texttt{forward\_train()} using the same hybrid system prompt system as PersonaPlex~\citep{roy2026personaplexvoicerolecontrol} and agent inner-monologue text format as Moshi~\citep{defossez2024moshi}. The text system prompt itself is unchanged from the PersonaPlex default for unstructured conversations (\texttt{"<system> You enjoy having a good conversation. <system>"}).
The voice prompt and inner-monologue text is pre-processed for the Seamless dataset as follows:

\paragraph{Voice system prompt.}
PersonaPlex's hybrid prompt requires a $\sim$10\,s pre-tokenized voice
clip per agent (8 Mimi codebooks) to prime the model
on the agent's voice identity. For each Seamless agent we use a clean
$\sim$10\,s voice sample of that speaker, Mimi-tokenized in advance.
Roughly $1.4\%$ of pairs are dropped because no voice clip is available
for one of the two participants.
\paragraph{Agent inner-monologue text stream.}
PersonaPlex predicts a text stream along with voice stream at matching 12.5Hz token rate. The agent side of the word-aligned seamless conversation transcripts are preprocessed to the dense token format as in PersonPlex and Moshi. In this format, sub-word tokens for a word are temporally aligned to the beginning of the word's utterance in the audio stream, and remaining frames are filled with PAD and EPAD tokens.

\subsection{Additional Details of Motion Tower}
\label{appendix:motion_tower_training}
This subsection covers the training-time hyperparameters of the motion
tower; the architecture and loss are described in
\cref{sec:method}. 
\paragraph{Training details.}
We use AdamW~\cite{loshchilov2017decoupled} optimizer with
$\beta_1 = 0.9$, $\beta_2 = 0.95$, weight decay $0.1$, gradient clipping
at $\ell_2$-norm $1.0$, and learning rate $3 \times 10^{-4}$, effective batch size 512, and train our model on $64$ NVIDIA H100 GPUs for 30K iterations.  

\paragraph{Inference.}
We employ streaming autoregressive sampling with temperature $\tau = 1.0$
and no top-$k$ truncation.

\subsection{Details of causal RVQ-VAE}
\label{appendix:rvq_details}

We adapt the part-aware RVQ-VAE of GestureLSM~\citep{liu2025gesturelsm}
into a streaming codec for Seamless Interaction~\citep{seamless2025}.
This appendix covers per-part input dimensions, the streaming
architecture changes, and how the four decoded outputs assemble into a
complete SMPL-X pose at inference.

\paragraph{Per-part decomposition.}
Seamless body annotations follow the SMPL-H convention; we extract
SMPL-X~\cite{SMPL-X:2019}body+face
parameters by joint-set re-mapping and combine the result with
SPECTRE~\cite{filntisis2022visual} -extracted
FLAME~\cite{FLAME:SiggraphAsia2017}face parameters. The
four part-aware RVQ-VAEs operate on disjoint slices of this combined
representation:

\begin{table}[h]
  \centering
  \small
  \setlength{\tabcolsep}{4pt}  
  \renewcommand{\arraystretch}{1.15} 
  \begin{tabular}{@{} l c l c c @{}} 
    \toprule
    Part & \makecell{Input \\ dim} & Joints / fields & \makecell{$K$ \\ quantizers} & \makecell{Codebook \\ band} \\
    \midrule
    Upper & 78 & 13 joints $\{3, 6, 9, 12{-}21\} \times$ 6D rot & $K_b=6$ & $[0, 1024)$ \\
    Hands & 180 & 30 joints $\{25{-}54\} \times$ 6D rot & $K_b=6$ & $[1024, 2048)$ \\
    \makecell[l]{Lower \\ + trans} & 57 & \makecell[l]{9 joints $\{0,1,2,4,5,7,8,10,11\} \times$ 6D rot \\ $+$ 3D translation velocity} & $K_b=6$ & $[2048, 3072)$ \\
    Face & 56 & \makecell[l]{6D jaw rot (joint 22) \\ $+$ 50D FLAME expression} & $K_f=4$ & $[3072, 4096)$ \\
    \bottomrule
  \end{tabular}
\end{table}

Each part is independent at the codec level (separate encoder, decoder,
codebook); the four streams are unified only at the motion-tower input
where their token IDs are interleaved into the dyadic stream described
in \cref{sec:motion_tower}. The shared $V_\text{mot}{=}4096$ vocabulary
is the union of the four disjoint $1024$-entry bands.

\paragraph{Architecture modifications.}
We make only two changes to the public GestureLSM RVQ-VAE; every other
encoder, decoder, and quantizer setting is inherited unchanged.
\textbf{(i)} We make the encoder \emph{causal} by padding each dilated
convolution on the left only, so the code emitted at frame $f$ depends on
input frames up to $f$ and never on future frames---giving zero
look-ahead at deployment. \textbf{(ii)} We halve the temporal
downsampling, from the original $4{\times}$ to $2{\times}$, using a single
strided downsampling stage instead of two. Applied to motion at the
$25$~fps rate used for tokenization (below), this $2{\times}$ factor
yields a $12.5$~Hz token stream that aligns one-to-one with the
$12.5$~Hz speech features.

\paragraph{Training data.}
The four part-codecs are trained on Seamless Interaction body motion that
has passed through the preprocessing of \cref{appendix:dataset_filtering}:
the clip-level aspect-ratio filter and the per-frame validity mask,
translation rescaling and Savitzky--Golay smoothing, a fixed $180^\circ$
rotation about the $x$-axis that maps the recovered poses into the SMPL-X
world frame, and conversion to a $6$D-rotation pose representation. The
face codec is trained on the SPECTRE-extracted FLAME parameters; pairs in
which either speaker has $\geq 20\%$ frames with failed face detection are
excluded from face-codec training only.

\paragraph{Training recipe.}
Each part-codec is trained independently on a single NVIDIA H100 GPU,
using the same optimizer and reconstruction objective as the public
GestureLSM model, and is selected by held-out reconstruction error.

\paragraph{Tokenization.}
Body and face motion are resampled to $25$~fps and encoded; with the
$2{\times}$ downsampling this produces token streams at $f_m{=}12.5$~Hz.
For motion-tower training the encoders are run once per training pair
offline and the codes cached to disk as $16$-bit integer bins of shape
$(T, 22)$, so the training loop never re-invokes the codec. At inference
the encoders run online on the partner's observed motion to produce the
partner token stream that is teacher-forced into
\cref{eq:agent_sampling}, while the agent's own tokens are sampled from
the motion tower and decoded by the four part-decoders.

\paragraph{Output assembly and SMPL-X forward kinematics.}
Each decoder reconstructs the SMPL-X pose parameters of its joint subset.
The four reconstructions are concatenated in canonical SMPL-X joint order
(global orientation $+$ $54$ body / hand / face joints $+$ translation
$+$ FLAME expression) and passed through the neutral SMPL-X model under forward kinematics
with translation set to zero, matching the zero-translation evaluation
protocol used for all body baselines. The output is per-frame body+face
joint positions and mesh vertices.

\section{Details of Evaluations}

\subsection{Details of Metrics}
All metrics are computed by our own evaluation pipeline; for the
distributional metrics we reuse the reference Fr\'echet-distance
implementation from ViBES~\citep{zhang2026vibes}.

\paragraph{FGD}
Fr\'echet distance on the agent's raw $22\times3=66$-D SMPL-X joint
positions per frame, pooled across the test set. We adopt the FGD term
from \citet{yoon2020trimodal}, but compute the Fr\'echet distance
directly on raw joint positions (no learned feature extractor),
following the raw-geometry protocol of recent dyadic-gesture work
(SARAH~\citep{ng2026sarah}, DyaDiT~\citep{peng2026dyadit}) rather than
the learned autoencoder latent space of the original FGD.

\paragraph{BeatAlign}
Gaussian-Average Hit Rate between the agent's audio onsets and
per-joint motion velocity-minima on the 13 upper-body joints,
following the AIST++~\citep{aichoreo} / BEAT~\citep{liu2022beat}
formulation with the audio-anchored alignment direction of
ViBES~\citep{zhang2026vibes} (for each audio onset, the distance to
the nearest motion beat); audio onsets are extracted at Mimi's
$24$\,kHz native sample rate.

\paragraph{Diversity}
Mean L2 distance between random pairs of agent-pose frames pooled
across the test set, following the Audio2Photoreal~\citep{ng2024audio2photoreal}
formulation (their \texttt{calculate\_diversity}); closer to GT is better.

\paragraph{P-FD}
Paired Fr\'echet
Distance~\citep{ng2022learning2listen, maluleke2024synergy}, computed
on $132$-D per-frame vectors formed by concatenating both speakers'
$22\times3$ joint positions, pooled across the test set. We report
the Fr\'echet distance between $(\text{GT}_A, \text{GT}_B)$ and
$(\text{GT}_A, \text{Gen}_B)$ distributions.

\paragraph{$\Delta$-User}
Relative gain in P-FD when each pair's user-motion track is replaced
by a shuffled (mismatched) one; see main text for the formula.

\subsection{DualTalk Implementation Details}
We adapt DualTalk~\cite{peng2025dualtalk} from its original face-blendshape
regression setting to SMPL-X body-pose regression on the Seamless
Interaction dataset. We keep DualTalk's four-module architecture---(a)
Dual-Speaker Joint Encoder, (b) Cross-Modal Temporal Enhancer, (c)
Dual-Speaker Interaction Module, and (d) Expressive Synthesis Module---at
its original depth (2 LSTM layers, 3 transformer encoder layers, 1
transformer decoder layer), together with the Wav2Vec~2.0 large audio
backbone (\texttt{facebook/wav2vec2-large-960h-lv60-self}).\footnote{DualTalk
public release: \url{https://github.com/ziqiaopeng/DualTalk}\label{fn:dualtalk}}
We make the following changes.
\paragraph{Output representation.}
We replace DualTalk's 56-D facial blendshape output (50 expression $+$ 3
jaw $+$ 3 neck) with a 69-D SMPL-X body parameterization (63-D body pose
$+$ 3-D global orientation $+$ 3-D translation, all in axis-angle). The
input motion encoder and the output projection head are resized
accordingly.
\paragraph{Training.}
Each speaker's audio is encoded by a Wav2Vec~2.0 backbone, initialized
from public pre-trained weights, with the convolutional feature extractor
frozen and the remaining Wav2Vec~2.0 layers fine-tuned; all other
components---the audio projection, the motion encoder, and modules
(b)--(d)---are randomly initialized and trained. We use DualTalk's
original objective, a sum of per-component mean-squared errors plus a
frame-to-frame velocity term.\textsuperscript{\ref{fn:dualtalk}} We
optimize with Adam at a learning rate of $1{\times}10^{-4}$ on 8 GPUs. The
Seamless Interaction dataset ($4{,}000$\,h) is roughly $80\times$ larger
than the 50-hour dataset used in the original DualTalk
work.\textsuperscript{\ref{fn:dualtalk}}
\paragraph{Discussion.}
Trained on Seamless body data, DualTalk collapses to a near-constant body
pose (per-frame body-velocity std $<\!10^{-5}$, vs.\ $\sim\!10^{-2}$ for
our model), with the velocity term decreasing throughout training as the
output converges to a static pose. We attribute this to two factors. (1)
Under a mean-squared-error objective, the sparse, low-amplitude gestures
in conversational body motion make the constant mean-pose solution a
strong local optimum. (2) DualTalk's architecture and loss were designed
for audio-to-face (lip-sync) generation, where the audio-to-motion mapping
is near-deterministic and temporally dense; audio-to-body gesture is only
weakly and non-deterministically coupled to speech, providing little
reliable per-frame signal to learn---independent of how the model is
initialized.

\subsection{Audio2Photoreal Implementation Details}

We adapt Audio2Photoreal~\cite{ng2024audio2photoreal}, originally a full-body photorealistic-avatar synthesizer trained on Meta's proprietary high-fidelity dyadic capture, to our body-pose evaluation setting on the publicly released Seamless Interaction dataset. We keep the three-stage architecture---(a) a TemporalVertexCodec VQ-VAE that tokenizes body motion into a discrete codebook, (b) a Guide transformer that auto-regressively predicts keyframe tokens conditioned on audio, and (c) a FiLMTransformer diffusion model that denoises the dense per-frame motion conditioned on audio plus the Guide's keyframe tokens---and the public-release classifier-free guidance recipe\footnote{Audio2Photoreal public release: 
 \url{https://github.com/facebookresearch/audio2photoreal/}\label{fn:a2p}}. We make the following changes.

\paragraph{Output representation.}
A2P natively predicts latent face-expression codes and kinematic-skeleton body joint angles; we replace this output with a 132-D SMPL-X body parameterization (22 joints in 6-D continuous rotation~\cite{zhou2019continuity}). We subtract each clip's mean translation and run SMPL-X forward kinematics with zero global translation (pelvis at the origin), matching the zero-translation evaluation protocol. The model generates agent~B's body from the conversational audio of both speakers; A2P is audio-driven and takes no motion input. Because our task is body-only, we discard the public release's face branch and retrain only the body pathway.

\paragraph{Architecture and training objective.}
We use A2P's three-stage architecture, model capacities, training losses, and optimizer settings exactly as released\textsuperscript{\ref{fn:a2p}}, including the frozen vq-wav2vec audio encoder. The only model-side change is for frame rate: the public code hard-codes the audio-feature length and conditioning positions for 30\,fps motion, so at our 25\,fps rate we recompute them once from the (deterministic, fixed-stride) wav2vec feature extractor for the target window length.

\paragraph{Data and training schedule.}
We train on the Seamless Interaction subset shared by our other baselines, using 24-second windows at 25\,fps. The tokenizer (\num{300000} iterations) and Guide (\num{30000} steps) are each trained on a single GPU; for the diffusion model we depart from the single-GPU public recipe and train across 8 GPUs (per-GPU batch~4, an effective batch of~32) for \num{800000} steps.

\paragraph{Inference.}
Inference follows the public recipe---chunked diffusion over consecutive 24-second windows with the released classifier-free guidance---after which we recover meshes by SMPL-X forward kinematics under the zero-translation convention above. 

\subsection{Details of User Study}
\begin{figure}[h!]
  \centering
  \includegraphics[width=0.6 \linewidth]{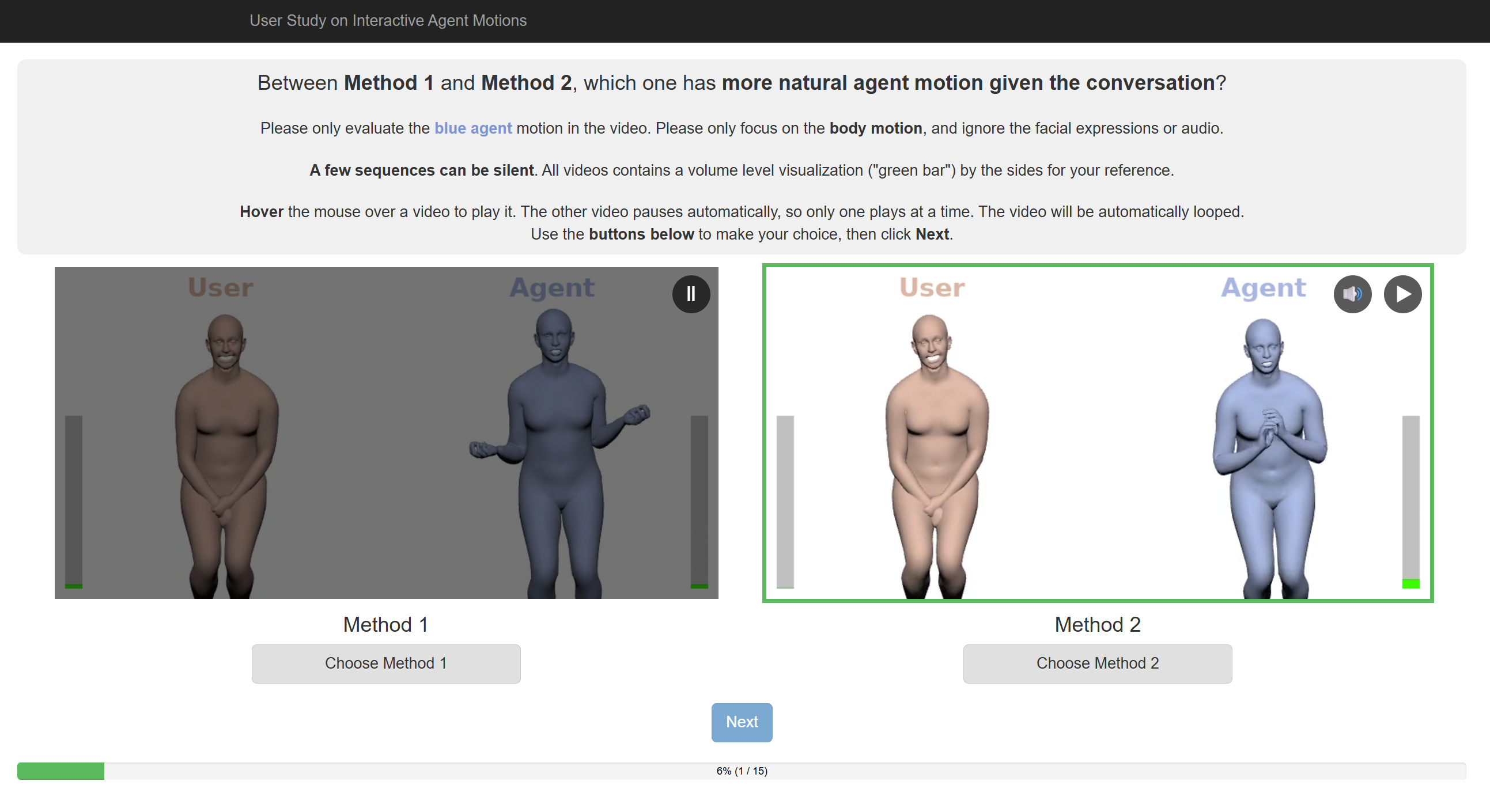}
  \caption{Interactive interface for the user study.}
  \label{fig:user_study}
\end{figure}
To evaluate the perceived quality of the generated body motion, we designed a web-based interactive interface to playback and compare the result videos. An example is shown in Figure~\ref{fig:user_study}. 
Note that the two videos can be played independently so the participants can freely toggle between the two methods for evaluation and comparisons.
In each result video, we render the animated mesh rendering of the partner (in peach color) and the generated agent (in blue color) along with the conversation audios. We randomly sample the comparison type (vs. the ground-truth, vs. Audio2Photoreal, vs. DualTalk) and randomly choose the display order on the web interface within the pair. Each pair of result videos shows the identical input partner and audio, with generated agent motion from different methods.
The partner audio is encoded to the left audio channel and the generated agent audio is encoded to the right channel. Consequently, we require all participants to wear a headphone that supports stereo sound for this study.
Since some input sequences contain near silent audio, we further visualize the live audio volume on the result videos, as a reference for the participants.
During the evaluation, we ask the participants to focus on the body motion, and ignore the facial expressions or audio.